\pdfoutput=1

\documentclass[11pt]{article}

\usepackage{acl}

\usepackage{times}
\usepackage{latexsym}

\usepackage[T1]{fontenc}

\usepackage[utf8]{inputenc}

\usepackage{microtype}
\usepackage{graphicx}
\usepackage{bm}
\usepackage{amssymb}
\usepackage{amsmath}
\usepackage{multirow}
\usepackage{booktabs}

\usepackage{hyperref}
\hypersetup{hidelinks,
    colorlinks=true,
    allcolors=black,
    pdfstartview=Fit,
    breaklinks=true}

\title{ERNIE-Layout: Layout Knowledge Enhanced Pre-training  \\ for Visually-rich Document Understanding}

\author{
\textbf{Qiming Peng$^{1}$\thanks{~ Equal contribution.}$\;$, Yinxu Pan$^{1}$\footnotemark[1]$\;$, Wenjin Wang$^{2}$\footnotemark[1]$\;$\thanks{~ Work done during internship at Baidu Inc.}$\;$, Bin Luo$^{1}$\thanks{~ Corresponding author: Bin Luo.}$\;$, Zhenyu Zhang$^{1}$,} \\ 
\textbf{Zhengjie Huang$^{1}$, Teng Hu$^{1}$, Weichong Yin$^{1}$, Yongfeng Chen$^{1}$, Yin Zhang$^{2}$,} \\ 
\textbf{Shikun Feng$^{1}$, Yu Sun$^{1}$, Hao Tian$^{1}$, Hua Wu$^{1}$, Haifeng Wang$^{1}$} \\
$^{1}$Baidu Inc., Beijing, China \\
$^{2}$Zhejiang University, Hangzhou, China \\
{\small \texttt{\{pengqiming, panyinxu, luobin06, zhangzhenyu07\}@baidu.com}} \\
{\small \texttt{\{wangwenjin, zhangyin98\}@zju.edu.cn} \quad \texttt{\{yinweichong, fengshikun, sunyu02\}@baidu.com}}
}

\begin{document}
\maketitle
\begin{abstract}

Recent years have witnessed the rise and success of pre-training techniques in visually-rich document understanding.
However, most existing methods lack the systematic mining and utilization of layout-centered knowledge, leading to sub-optimal performances.
In this paper, we propose ERNIE-Layout, a novel document pre-training solution with layout knowledge enhancement in the whole workflow, to learn better representations that combine the features from text, layout, and image.
Specifically, we first rearrange input sequences in the serialization stage, and then present a correlative pre-training task, reading order prediction, to learn the proper reading order of documents.
To improve the layout awareness of the model, we integrate a spatial-aware disentangled attention into the multi-modal transformer and a replaced regions prediction task into the pre-training phase. 
Experimental results show that ERNIE-Layout achieves superior performance on various downstream tasks, setting new state-of-the-art on key information extraction, document image classification, and document question answering datasets. 
The code and models are publicly available at PaddleNLP\footnote{\url{https://github.com/PaddlePaddle/PaddleNLP/tree/develop/model_zoo/ernie-layout}}.

\end{abstract}

\section{Introduction}
Visually-rich Document Understanding (VrDU) is an important research field aiming to handle various types of scanned or digital-born business documents (e.g., forms, invoices), which has attracted great attention from the industry and academia due to its various applications.
Distinct from conventional natural language understanding (NLU) tasks that use only plain text, VrDU models have the opportunity to access the most primitive data features.
Herein, the diversity and complexity of document formats pose new challenges to the task, an ideal model needs to make full use of the textual, layout, and even visual information to fully understand visually-rich document like humans.

The preliminary works for VrDU~\cite{yang2016hierarchical,yang2017learning,katti2018chargrid,sarkhel2019deterministic,cheng2020one} usually adopt uni-modal or shallow multi-modal fusion approaches, which are task-specific and require massive data annotations.
Recently, pre-training language models have swept the field, LayoutLM~\cite{xu2020layoutlm}, LayoutLMv2~\cite{xu2021layoutlmv2}, and some advanced document pre-training approaches~\cite{li2021structurallm,appalaraju2021docformer,gu2022xylayoutlm} have been proposed successively and achieved great successes in various VrDU tasks.
Unlike popular uni-modal or vision-language frameworks~\cite{devlin2019bert,liu2019roberta,lu2019vilbert,yu2021ernie}, the uniqueness of document understanding models lies in how to exploit the layout knowledge. 

However, existing document pre-training solutions typically fall into the trap of simply taking 2D coordinates as an extension of 1D positions to endow the model layout awareness. 
Considering the characteristics of VrDU, we believe that the layout-centered knowledge should be systematically mined and utilized from two aspects:
(1) On the one hand, \emph{layout implicitly reflects the proper reading order of documents}, while previous methods are used to perform the serialization by multiplexing the results of Optical Character Recognition (OCR), which roughly arrange tokens in the top-to-bottom and left-to-right manner~\cite{wang2021layoutreader,gu2022xylayoutlm}. 
Inevitably, it is inconsistent with human reading habits for documents with complex layouts (e.g., tables, forms, multi-column templates) and leads to sub-optimal performances for downstream tasks.
(2) On the other hand, \emph{layout is actually the third modality besides language and vision}, while current models are used to take layout as a special position feature, such as the layout embedding in input layer~\cite{xu2020layoutlm} or the bias item in attention layer~\cite{xu2021layoutlmv2}.
The lack of cross-modal interaction between layout and text/image might restrict the model from learning the role of layout in semantic expression.

To achieve these goals, we propose a systematic layout knowledge enhanced pre-training approach, ERNIE-Layout\footnote{It is named after the knowledge enhanced pre-training model, ERNIE~\cite{sun2019ernie}, as a layout enhanced version.}, to improve the performances of document understanding tasks. 
First of all, we employ an off-the-shelf layout-based document parser in the serialization stage to generate an appropriate reading order for each input document, so that the input sequences received by the model are more in line with human reading habits than using the rough raster-scanning order.
Then, each textual/visual token is equipped with its position embedding and layout embedding, and sent to the stacked multi-modal transformer layers.
To enhance cross-modal interaction, we present a spatial-aware disentangled attention mechanism, inspired by the disentangled attention of DeBERTa~\cite{he2021deberta}, in which the attention weights between tokens are computed using disentangled matrices based on their hidden states and relative positions. 
In the end, layout not only acts as the 2D position attribute of input tokens, but also contributes a spatial perspective to the calculation of semantic similarity.

With satisfactory serialization results, we propose the pre-training task, reading order prediction, to predict the next token for each position, which facilitates the consistency within the same arranged text segment and the discrimination between different segments.
Furthermore, when pre-training, we also adopt the classic masked visual-language modeling and text-image alignment tasks~\cite{xu2021layoutlmv2}, and present a fine-grained multi-modal task, replaced regions prediction, to learn the correlation among language, vision and layout. 

We construct broad experiments on three representative VrDU downstream tasks with six publicly available datasets to evaluate the performance of the pre-trained model, i.e., the key information extraction task with the FUNSD~\cite{jain2019multimodal}, CORD~\cite{park2019cord}, SROIE~\cite{huang2019icdar2019}, Kleister-NDA~\cite{gralinski2021kleister} datasets, the document question answering task with the DocVQA~\cite{mathew2021docvqa} dataset, and the document image classification task with the RVL-CDIP~\cite{harley2015evaluation} dataset.
The results show that ERNIE-Layout significantly outperforms strong baselines on almost all tasks,  proving the effectiveness of our two-part layout knowledge enhancement philosophy.

The contributions are summarized as follows:
\begin{itemize}
    \item ERNIE-Layout proposes to rearrange the order of input tokens in serialization and adopt a reading order prediction task in pre-training. To the best of our knowledge, ERNIE-Layout is the first attempt to consider the proper reading order in document pre-training.  
    \item ERNIE-Layout incorporates the spatial-aware disentangled attention mechanism in the multi-modal transformer, and designs a replaced regions prediction pre-training task, to facilitate the fine-grained interaction across textual, visual, and layout modalities.
    \item ERNIE-Layout refreshes the state-of-the-art of various VrDU tasks, and extensive experiments demonstrate the effectiveness of exploiting layout-centered knowledge.
\end{itemize}

\begin{figure*}
    \centering
    \includegraphics[width=1\textwidth]{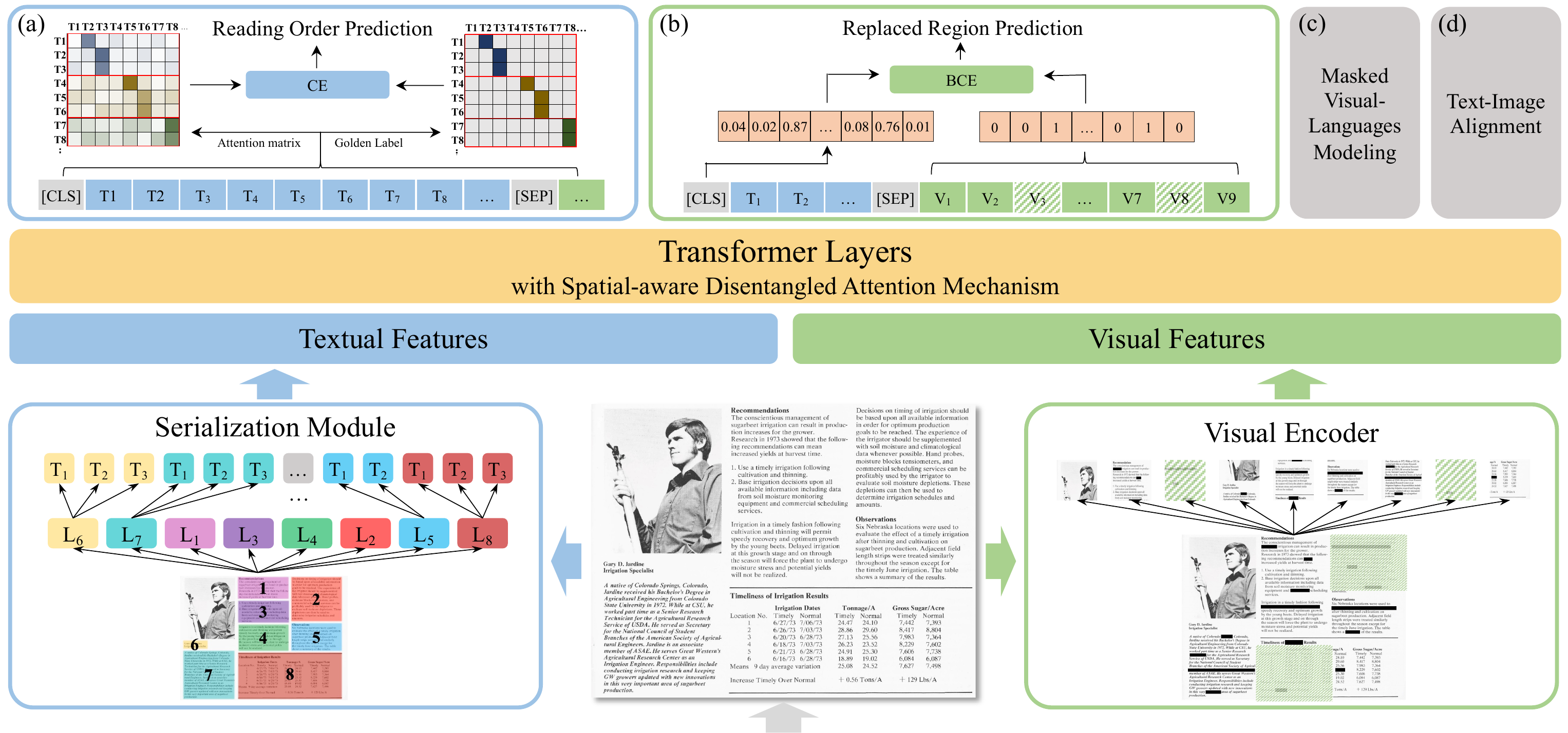}
    \caption{The architecture and pre-training objectives of ERNIE-Layout. The serialization module is introduced to correct the order of raster-scan, and the visual encoder extracts corresponding image features. With the spatial-aware disentangled attention mechanism, ERNIE-Layout is pre-trained with four tasks.}
    \label{fig:model}
\end{figure*}

\section{Related Work}

\noindent\textbf{Layout-aware Pre-trained Model.}
Humans understand visually rich documents through many perspectives, such as language, vision, and layout.
Based on the powerful modeling ability of Transformer~\cite{vaswani2017attention}, 
LayoutLM~\cite{xu2020layoutlm} initially embeds the 2D coordinates as layout embeddings for each token and extends the famous masked language modeling pre-training task~\cite{devlin2019bert} to masked visual-language modeling, which opens the prologue of layout-aware pre-trained models.
Afterwards, LayoutLMv2~\cite{xu2021layoutlmv2} concatenates document image patches with textual tokens, and two pre-training tasks, text-image matching and text-image alignment, are proposed to realize the cross-modal interaction.
StructralLM~\cite{li2021structurallm} leverages segment-level, instead of word-level, layout features to make the model aware of which words come from the same cell.
DocFormer~\cite{appalaraju2021docformer} shares the learned spatial embeddings across modalities, making it easy for the model to correlate text to visual tokens and vice versa.
TILT~\cite{powalski2021going} proposes an encoder-decoder model to generate results that are not explicitly included in the input sequence to solve the limitations of sequence labeling.
However, these methods ignore the potential value of layout in-depth and directly rely on a raster-scanning serialization, which is contrary to human reading habits.
To solve this problem, LayoutReader~\cite{wang2021layoutreader} designs a sequence-to-sequence framework to generate an appropriate reading order for each document. Unfortunately, it is carefully designed for reading order detection and cannot directly empower various document understanding tasks.
Besides, the above methods are used to regard layout as a subsidiary feature of text along with the idea of LayoutLM, but the same text with different layouts may also express different semantics. Therefore, we believe that layout should be regarded as the third modality independent of language and vision.

\noindent\textbf{Knowledge-enhanced Representation.}
Following the BERT~\cite{devlin2019bert} architecture, many efforts are devoted to pre-trained language models for learning informative representations. 
There are some studies show that extra knowledge, such as facts in WikiData and WordNet, can further benefit the pre-trained models~\cite{zhang2019ernie,liu2020kbert,he2020bert,wang2021kepler}, but the embeddings of words in the text and entities in the knowledge graphs are not in the same vector space, so a cumbersome adaptation module is required~\cite{he2020bert,wang2021kadapter}.
Another research line is to excavate the potential human cognitive laws of the text itself:
ERNIE~\cite{sun2019ernie} creativity proposes entity-level mask in pre-training to incorporate the human knowledge into language models.
Similarly, SpanBERT~\cite{joshi2020spanbert} modifies the making schema and training objectives to better represent and predict text spans.
BERT-wwm~\cite{cui2021pre} introduces a whole word masking strategy for Chinese language models.
Outside the field of plain text, ERNIE-ViL~\cite{yu2021ernie} incorporates structured knowledge obtained from scene graphs to learn joint representations of vision-language.
Inspired by the above work, we leverage the implicit knowledge related to layout, e.g., reading order, for the understanding of visually rich documents.

\section{Methodology}

Figure~\ref{fig:model} shows an overview of the ERNIE-Layout.
Given a document, ERNIE-Layout rearranges the token sequence with the layout knowledge and extracts visual features from the visual encoder.
The textual and layout embeddings are combined into textual features through a linear projection, and similar operations are executed for visual embeddings. 
Next, these features are concatenated and fed into the stacked multi-modal transformer layers, which are equipped with the proposed spatial-aware disentangled attention mechanism.
For pre-training, ERNIE-Layout adopts four pre-training tasks, including the new proposed reading order prediction, replaced region prediction tasks, and the traditional masked visual-language modeling, text-image alignment tasks.

\subsection{Serialization Module}
\label{sec:document-parser}

\begin{figure}[t]
    \includegraphics[width=0.5\textwidth]{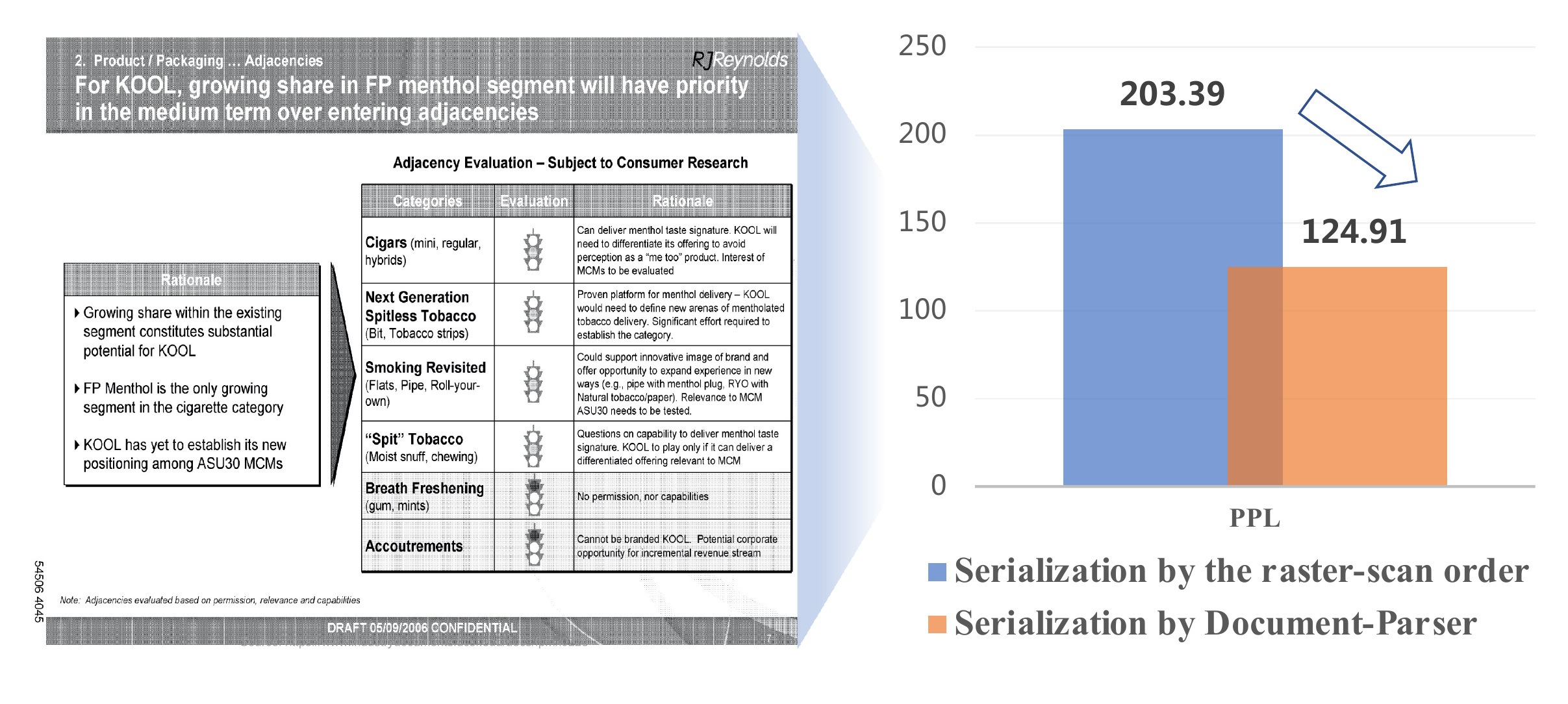}
    \caption{The effect of layout knowledge enhanced serialization compared with vanilla raster-scanning order. By using Document-Parser, the perplexity of the document with a complex layout is significantly reduced.}
    \label{fig:parser_example}
\end{figure}

Before feeding a visually rich document to neural networks, serialization, that is, recognizing the text and arranging them in proper order, is a necessary step.
First, an OCR tool is used to obtain the words and their coordinates in documents. Then the traditional method arranges the identified elements from left to right and top to bottom by raster-scan to generate the input sequence.
Although this method is easy to implement, it cannot correctly handle documents with complex layouts.
Look at the example in Figure~\ref{fig:parser_example}, there are two tables, and the cells in these tables also have some newline text. Suppose we want to extract some information from it. In that case, the expected results may not be obtained according to the raster-scanning order, because the words in the same cell are scattered.

Inspired by the human reading habits, we adopt Document-Parser, an advanced document layout analysis toolkit based on Layout-Parser\footnote{\url{https://github.com/Layout-Parser/layout-parser}}, to serialize these documents.
As shown in Figure~\ref{fig:model}, based on the words and their boxes recognized by OCR, it first detects document elements (e.g., paragraphs, lists, tables, figures), and then uses specific algorithms to obtain the logical relationship between words based on the characteristics of different elements, to obtain the proper reading order.

To quantitatively analyze the benefits of layout knowledge enhanced serialization, we take perplexity (PPL), calculated by GPT-2~\cite{radford2019language}, as the evaluation metric. PPL is widely used for measuring the performance of language models. 
From Figure~\ref{fig:parser_example}, we find that the input sequence serialized by Document-Parser has a lower PPL  than the raster-scanning order. More implementation details and cases are detailed in Appendix~\ref{sec:document_parser}.

\subsection{Input Representation}

The input sequence of ERNIE-Layout includes a textual part and a visual part, and the representation of each part is a combination of its modal features and layout embeddings~\cite{xu2021layoutlmv2}.

\noindent\textbf{Text Embedding.}
The document tokens after the serialization module are used as the text sequence. 
Following the pre-processing of BERT-Style models~\cite{devlin2019bert}, two special tokens \verb|[CLS]| and \verb|[SEP]| are appended at the beginning and end of the text sequence, respectively. 
Finally, the text embedding of token sequence $T$ is expressed as:
\begin{equation}
    \textbf{T}= \textbf{E}_{tk}(T) + \textbf{E}_{1p}(T) + \textbf{E}_{tp}(T),
\end{equation}
where $\textbf{E}_{tk}$, $\textbf{E}_{1p}$, $\textbf{E}_{tp}$ respectively denote the token embedding, 1D position embedding, and token type embedding layer.

\noindent\textbf{Visual Embedding.}
To extract the visual features of documents, we employ Faster-RCNN~\cite{ren2015faster} as the backbone of visual encoder.
In particular, the document image is resized to $224 \times 224$ and fed into visual backbone, an adaptive pooling layer is introduced to convert the output into a feature map with a fixed width $W$ and height $H$ (here, we set them to $7$).
Next, we flatten the feature map into a visual sequence $V$, and project each visual token to the same dimension as text embedding with a linear layer $F_{vs}(\cdot)$. 
Similarity, the 1D position and token type $\verb|[V]|$ are taken into consideration for the generation of visual embedding:
\begin{equation}
    \textbf{V}= F_{vs}(V) + \textbf{E}_{1p}(V) + \textbf{E}_{tp}(\verb|[V]|).
\end{equation}

\noindent\textbf{Layout Embedding.}
For each textual token, the OCR tool provides its 2D  coordinates with the width and height of the bounding box $(x_0, y_0, x_1, y_1, w, h)$,
where $(x_0, y_0)$ denote coordinates of the upper left corner of the bounding box, $(x_1, y_1)$ denote the bottom right corner, $w = x_1 - x_0$, $h = y_1 - y_0$, and all the coordinate values are normalized in the range $[0, 1000]$.
For the visual token, similar calculation processes can also be performed.
To look up the layout embeddings of textual/visual token, we construct separate embedding layers in the horizontal and vertical directions:
\begin{equation}
\textbf{L} = \textbf{E}_{2x}(x_0 ,x_1, w) + \textbf{E}_{2y}(y_0, y_1, h),
\end{equation}
where $\textbf{E}_{2x}$ is the x-axis embedding layer, $\textbf{E}_{2y}$ denotes the y-axis embedding layer.

To achieve the ultimate input representation $\textbf{H}$ of ERNIE-Layout,
we integrate the embedding of each textual and visual token with its corresponding layout embeddings.
Finally, the textual and visual embeddings are combined together to obtain a long sequence with the length being $N+HW$, where $N$ is the max length of the textual part:
\begin{equation}
\textbf{H} = [\textbf{T}+ \textbf{L};\textbf{V}+ \textbf{L}].
\end{equation}

\subsection{Multi-modal Transformer}
\label{sec:muti-modal-transformer}

In the final input representation, textual and visual tokens are spliced together, and the self-attention mechanism in the transformer supports their layer-aware cross-modal interaction.
However, as a unique modality, layout features should be involved in the calculation of attention weight, and the tightness between them and contents (collectively refers to text and image) should also be taken into account explicitly.
Inspired by the disentangled attention of DeBERTa~\cite{he2021deberta}, in which the attention weights among tokens are computed using disentangled matrices on their contents and relative positions, we propose spatial-aware disentangled attention for the multi-modal transformer to enable the participation of layout features.

Firstly, we take 1D position as an example to define the relative distance $\delta_{1p}(\cdot)$ between token $i$ and $j$, and the definition in the $x$-axis and $y$-axis directions of 2D layout is the same:
\begin{equation}
    \delta_{1p}(i, j) =\begin{cases}0 & {\rm for}\ \ i - j \leqslant -k\\2k-1 & {\rm for}\  \ i-j \geqslant k\\i-j+k & {\rm others,}\end{cases}
\end{equation}

Next, to construct relative position vectors consistent with the input dimension, we introduce three relative position embedding tables $\textbf{E}_{1p}^{'}$, $\textbf{E}_{2x}^{'}$, $\textbf{E}_{2y}^{'}$ for 1D position, 2D $x$-axis and 2D $y$-axis.
After looking up the embedding table, a series of projection matrices map these relative position vectors as well as the content vectors into $\textbf{Q}^{\star}$, $\textbf{K}^{\star}$, $\textbf{V}^{\star}$ in the attention mechanism, where $\star \in \{{ct, 1p, 2x, 2y}\}$.
In the process of attention calculation, we decouple the raw score into four parts to realize the in-depth exchange of 1D/2D features and contents:
\begin{align}
    {A}_{ij}^{ct,ct} &= \textbf{Q}^{ct}_{i} {\textbf{K}^{ct}_{j}}^{\top}, \\
    {A}_{ij}^{ct,1p} &= \textbf{Q}^{ct}_{i}{\textbf{K}^{1p}_{\delta_{1p}(i,j)}}^{\top} + \textbf{K}^{ct}_{j}{\textbf{Q}^{1p}_{\delta_{1p}(j,i)}}^{\top}, \\
    {A}_{ij}^{ct,2x} &= \textbf{Q}^{ct}_{i}{\textbf{K}^{2x}_{\delta_{2x}(i,j)}}^{\top} + \textbf{K}^{ct}_{j}{\textbf{Q}^{2x}_{\delta_{2x}(j,i)}}^{\top}, \\
    {A}_{ij}^{ct,2y} &= \textbf{Q}^{ct}_{i}{\textbf{K}^{2y}_{\delta_{2y}(i,j)}}^{\top} + \textbf{K}^{ct}_{j}{\textbf{Q}^{2y}_{\delta_{2y}(j,i)}}^{\top}.
\end{align}

Finally, all these attention scores are summed up to get the attention matrix $\hat{\textbf{A}}$. With the scaling and normalization operations, the output of spatial-aware disentangled attention is\footnote{The schematic workflow is shown in
Appendix~\ref{sec:detail_disentangle}}:
\begin{align}
    \hat{A}_{ij} &= {A}_{ij}^{ct,ct} + {A}_{ij}^{ct,1p} + {A}_{ij}^{ct,2x} + {A}_{ij}^{ct,2y}, \\
    \textbf{H}_{out} &= {\rm softmax}(\frac{\hat{\textbf{A}}}{\sqrt{3d}}) \textbf{V}^{ct}.
\end{align}

\subsection{Pre-training Tasks}

There are four pre-training tasks in ERNIE-Layout.
We design reading order prediction and replaced region prediction, as well as borrow masked visual-language modeling and text-image alignment from LayoutLMv2~\cite{xu2021layoutlmv2}, so that the model has the ability to learn layout knowledge and fuse various multi-modal information.

\noindent\textbf{Reading Order Prediction.}
The serialization result consists of several text segments, including a series of words and 2D coordinates. Based on the knowledge, we organize the input words in proper reading order. 
However, there is no explicit boundary between text segments in the input sequence received by the transformer. 
To make the model understand the relationship between layout knowledge and reading order and still work well when receiving input in inappropriate order, we propose Reading Order Prediction (ROP) and hope the attention matrix $\hat{\textbf{A}}$ carries the knowledge about reading order.
In this way, we give ${\hat{A}_{ij}}$ an additional meaning, i.e., the probability that the $j$-th token is the next token of the $i$-th token.
Besides, the ground truth is a 0-1 matrix $G$, where 1 indicates that there is a reading order relationship between the two tokens and vice versa. For the end position, the next token is itself.
In pre-training, we calculate the loss with Cross-Entropy:
\begin{equation}
    \mathcal{L}_{\rm ROP} = - \sum_{0 \leq i < N} \sum_{0 \leq j < N} G_{ij} \log(\hat{A}_{ij}).
\end{equation}

\noindent\textbf{Replaced Region Prediction.}
In visual encoder, each document image is processed into a sequence with a fixed length $HW$.
To enable the model perceive fine-grained correspondence between image patches and text, with the help of layout knowledge, we propose Replaced Region Prediction (RRP).
Specifically, $10\%$ of the patches are randomly selected and replaced with a patch from another image, the processed image is encoded by the visual encoder and input into the multi-modal transformer. 
Then, the \verb|[CLS]| vector output by the transformer is used to predict which patches are replaced. 
So the loss of this task is:
\begin{align}
    \mathcal{L}_{\rm RRP}= - \sum_{0 \leq i < HW} & [ G_i \log(P_i) + \nonumber \\ 
        & (1- G_i)\log(1- P_i)],
\end{align}
where $G_i$ is the golden label of replaced patches, $P_i$ is the normalized probability of prediction.

\noindent\textbf{Masked Visual-Language Modeling.}
Similar to masked language modeling (MLM), the objective of masked visual-language modeling (MVLM) is to recover the masked text token based on its text context and the whole multi-modal clues.

\noindent\textbf{Text-Image Alignment.} 
Besides the image-side cross-modal task RRP, we also adopt Text-Image Alignment (TIA), as a text-side task, to help the model learn the spatial correspondence between image regions and coordinates of bounding box.
Here, some text lines are randomly selected, and their corresponding regions are covered on the document image.
Then, a classification layer is introduced to predict whether each text token is covered.

To sum up, the final pre-training objective is:
\begin{equation}
\mathcal{L} = \mathcal{L}_{\rm ROP} + \mathcal{L}_{\rm RRP} + \mathcal{L}_{\rm MVLM} + \mathcal{L}_{\rm TIA}
\end{equation}

\section{Experiments}

\subsection{Datasets}

For the fairness of experiments, we only use \emph{layout knowledge enhanced serialization to rearrange the reading order of pre-training data}, which means that ERNIE-Layout receives the same input as the compared methods in the fine-tuning phase.

\noindent\textbf{Pre-training.}
Following popular choice in VrDU, we crawl the homologous data of IIT-CDIP Test Collection~\cite{lewis2006building} from Tabacco website, which contains over 30 million scanned document pages, and randomly select 10 million pages from them as the pre-training data. 

\noindent\textbf{Fine-tuning.}
We carry out broad experiments on various downstream VrDU tasks and datasets.
For the \emph{key information extraction} task, we select FUNSD~\cite{jain2019multimodal}, CORD~\cite{park2019cord}, SROIE~\cite{huang2019icdar2019}, and Kleister-NDA~\cite{gralinski2021kleister} as the evaluation datasets.
For the \emph{document question answering} task, the DocVQA~\cite{mathew2021docvqa} dataset is selected.
For the \emph{document image classification} task, we select the RVL-CDIP~\cite{harley2015evaluation} dataset.
Table \ref{tab:statistics_of_fine_tuning_datasets} shows the brief statistics of them and more details are included in Appendix \ref{sec:detail_finetuning_datasets}.

\begin{table}[t]
\centering
\small
\begin{tabular}{lrrrr}
\toprule
\textbf{Dataset} & \textbf{\#Field} & \textbf{\#Train} & \textbf{\#Dev} & \textbf{\#Test} \\
\midrule
FUNSD & 4 & 149 & - & 50 \\
CORD & 30 & 800 & 100 & 100 \\
SROIE & 4 & 626 & - & 347 \\ 
Kleister-NDA & 4 & 254 & 83 & 203 \\
RVL-CDIP & 16 & 320K & 40K & 40K \\
DocVQA & - & 39K & 5K & 5K \\ 
\bottomrule
\end{tabular}
\caption{Statistics of datasets for downstream tasks}
\label{tab:statistics_of_fine_tuning_datasets}
\end{table}

\begin{table}[t]
    \centering
    \small
    \begin{tabular}{lrrr}
    \toprule
    \textbf{Dataset} & \textbf{Epoch} & \textbf{Weight Decay} & \textbf{Batch} \\
    \midrule
    FUNSD & 100 & - & 2 \\
    CORD & 30 & 0.05 & 16 \\
    SROIE & 100 & 0.05 & 16 \\ 
    Kleister-NDA & 30 & 0.05 & 16 \\
    RVL-CDIP & 20 & 0.05 & 16 \\
    DocVQA & 6 & 0.05 & 16 \\ 
    \bottomrule
    \end{tabular}
    \caption{Hyper-parameters for downstream tasks}
    \label{tab:finetune_hyper_parameter}
\end{table}

\begin{table*}[t]
\centering
\small
\begin{tabular}{llcccccc}
\toprule
\# & \textbf{Methods} & \textbf{FUNSD} (F1) & \textbf{CORD} (F1) & \textbf{SROIE} (F1) & \textbf{Kleister-NDA} (F1) \\
\midrule
1 & BERT\textsubscript{large}~\cite{liu2019roberta}      & 0.6563 & 0.9025 & 0.9200 & 0.7910 \\
2 & RoBERTa\textsubscript{large}~\cite{liu2019roberta}   & 0.7072 & - & 0.9280 & -  \\
3 & UniLMv2\textsubscript{large}~\cite{bao2020unilmv2}   & 0.7257 & 0.9205 & 0.9488 & 0.8180  \\
\midrule
4 & LayoutLM\textsubscript{large}~\cite{xu2020layoutlm}            & 0.7895 & 0.9493 & 0.9524 & 0.8340 \\
5 & TILT\textsubscript{large}~\cite{powalski2021going}             & -      & 0.9633 & \textbf{0.9810} & - \\
6 & LayoutLMv2\textsubscript{large}~\cite{xu2021layoutlmv2}        & 0.8420 & 0.9601 & \underline{0.9781} & 0.8520 \\
7 & StructuralLM\textsubscript{large}~\cite{li2021structurallm}    & \underline{0.8514} & -      & -      & - \\
8 & DocFormer\textsubscript{large}~\cite{appalaraju2021docformer}  & 0.8455 & \underline{0.9699} & -      & \underline{0.8580} \\
\midrule
\midrule
9 & ERNIE-Layout\textsubscript{large} (ours)  & \textbf{0.9312} & \textbf{0.9721} & 0.9755 & \textbf{0.8810} \\ 
\bottomrule
\end{tabular}
\caption{Results (Entity-level F1 score) of ERNIE-Layout and previous methods on the \emph{Key Information Extraction} task (\emph{FUNSD}, \emph{CORD}, \emph{SROIE}, \emph{Kleister-NDA}). The highest and second-highest scores are bolded and underlined.}
\label{tab:main_result_ner}
\end{table*}

\begin{table*}[t]
\centering
\small
\begin{tabular}{llccc}
\toprule
\# &\textbf{Methods}             &\textbf{Fine-tuning set}       &\textbf{ANLS} &\textbf{${\vartriangle}$ANLS} \\
\midrule
1 & BERT\textsubscript{large}~\cite{liu2019roberta}     &train  &0.6768 &\\
2 & RoBERTa\textsubscript{large}~\cite{liu2019roberta}  &train  &0.6952 & \\
3 & UniLMv2\textsubscript{large}~\cite{bao2020unilmv2} &train &0.7709 & \\
\midrule
4 & LayoutLM\textsubscript{large}~\cite{xu2020layoutlm}         &train &0.7259 & \\
5 & TILT\textsubscript{large}~\cite{powalski2021going}          &-     &\textbf{0.8705} & \\
6 & StructuralLM\textsubscript{large}~\cite{li2021structurallm} &-     &0.8349 & \\
\midrule
7a & LayoutLMv2\textsubscript{large}~\cite{xu2021layoutlmv2} & train & 0.8348 & + 0.0639 (\#3) \\
7b & LayoutLMv2\textsubscript{large}~\cite{xu2021layoutlmv2}             & train + dev & \underline{0.8529} & + 0.0820 (\#3) \\
\midrule
\midrule
8a & ERNIE-Layout\textsubscript{large} (ours)   &train     &0.8321 & + \underline{0.1369} (\#2)\\ 
8b & ERNIE-Layout\textsubscript{large} (ours)   &train+dev &0.8486 & + \textbf{0.1534} (\#2) \\ 
\midrule
9 & ERNIE-Layout\textsubscript{large}~(leaderboard) &train+dev &\textbf{0.8841} & \\
\bottomrule
\end{tabular}
\caption{Results (Average Normalized Levenshtein Similarity, ANLS) of ERNIE-Layout and previous methods on the \emph{Document Question Answering} task (\emph{DocVQA}). "-" means the fine-tuning set is not clearly described in the original paper. ${\vartriangle}$ANLS means ANLS difference between the multi-modal model and its corresponding text-only model, where ERNIE-Layout is initialized from RoBERTa and LayoutLMv2 is initialized from UniLMv2.} 
\label{tab:main_result_docvqa}
\end{table*}

\subsection{Settings}

\noindent\textbf{Pre-training.}
ERNIE-Layout has 24 transformer layers with 1024 hidden units and 16 attention heads. 
The maximum sequence length of textual tokens is 512 the sequence length of visual tokens is 49.
The transformer is initialized from RoBERTa-large~\cite{liu2019roberta}, and the visual encoder takes Faster-RCNN~\cite{ren2015faster} as the initialized model. 
The rest parameters are randomly initialized.
We use Adam~\cite{kingma2014adam} as the optimizer, with a learning rate of 1e-4 and a weight decay of 0.01. The learning rate is linearly warmed up over the first 10\% steps, then linearly decayed to 0. 
ERNIE-Layout is trained on 24 Tesla A100 GPUs for 20 epochs with a batch size of 576. 

\noindent\textbf{Fine-tuning.}
We solve the \emph{key information extraction} tasks (FUNSD, CORD, SROIE, Kleister-NDA) with a sequence labeling framework and introduce a token-level classification layer to predict the BIO labels. 
For the \emph{document question answering} task (DocVQA), we follow the extractive question-answering paradigm and build a token-level classifier after the ERNIE-Layout output representation to predict the start and end position of the answer.
For the \emph{document image classification} task (RVL-CDIP), the representation of \verb|[CLS]| is processed by a fully-connected network to predict the document label. 
ERNIE-Layout is fine-tuned for all the downstream tasks using Adam optimizer, with a learning rate of 2e-5, weight decay of 0.01. 
Similar to pre-training, the learning rate is linearly warmed up and then linearly decayed.
Other hyper-parameters are shown in Table \ref{tab:finetune_hyper_parameter}.

\subsection{Results}
\label{sec:exp_results}

\noindent\textbf{Key Information Extraction.}
Table~\ref{tab:main_result_ner} shows the results on four datasets, in which we utilize entity-level F1 score to evaluate these sequence labeling tasks.
ERNIE-Layout achieves new state-of-the-art on FUNSD, CORD, Kleister-NDA, and competitive performance on SROIE.
It is worth mentioning that, in the FUNSD, ERNIE-Layout obtains a significant and stable improvement of 7.98\% (with a standard deviation 0.0011), compared to the previous best results.
The above phenomena are enough to verify the effectiveness of our design philosophy that mining and utilizing layout knowledge in document pre-training models.

\begin{table}[t]
\centering
\small
\begin{tabular}{llccccc}
\toprule
\# & \textbf{Methods}                    & \textbf{Accuracy} \\
\midrule
1 & BERT\textsubscript{large} \cite{liu2019roberta} & 0.8992 \\
2 & RoBERTa\textsubscript{large} \cite{liu2019roberta} & 0.9011 \\
3 & UniLMv2\textsubscript{large} \cite{bao2020unilmv2} & 0.9020 \\
\midrule
4 & LayoutLM\textsubscript{large} \cite{xu2020layoutlm} & 0.9443 \\
5 & TILT\textsubscript{large} \cite{powalski2021going}  &  0.9552 \\
6 & LayoutLMv2\textsubscript{large} \cite{xu2021layoutlmv2} & 0.9564 \\
7 & StructuralLM\textsubscript{large} \cite{li2021structurallm} & \underline{0.9608} \\
8 & DocFormer\textsubscript{large}~\cite{appalaraju2021docformer} & 0.9550 \\
\midrule
\midrule
9 & ERNIE-Layout\textsubscript{large}~(ours) & \textbf{0.9627} \\ 
\bottomrule
\end{tabular}
\caption{Results (Accuracy) of ERNIE-Layout and previous methods on the \emph{Document Image Classification} task (\emph{RVL-CDIP}).}
\label{tab:main_result_document_classification}
\end{table}

\begin{table*}[ht]
\centering
\small
\begin{tabular}{lccccccccc}
\toprule
\# &\textbf{MVLM} &\textbf{TIA} &\textbf{RRP} &\textbf{ROP} & \textbf{SADA} & \textbf{SASA} & \textbf{FUNSD} & \textbf{CORD} \\
\midrule
1   & $\surd$ &  &  &  &  &  & 0.8712 &0.9513 \\
2   & $\surd$ & $\surd$ &  &  &  &  & 0.8753 &0.9555\\
\midrule
3$^{\dag}$   & $\surd$ & $\surd$ & $\surd$ &  &  &  & 0.8848 & 0.9565\\
4$^{\dag}$   & $\surd$ & $\surd$ & $\surd$ & $\surd$ &  &  & 0.8978 &0.9603\\
5$^{\dag}$   & $\surd$ & $\surd$ & $\surd$ & $\surd$ & $\surd$ & & 0.9241 &0.9673\\
\midrule
6   & $\surd$ & $\surd$ & $\surd$ & $\surd$ & & $\surd$ & 0.9128 &0.9658\\
\bottomrule
\end{tabular}
\caption{Performance analysis with different pre-training tasks and attention mechanisms, in which SADA refers to the spatial-aware disentangled attention in ERNIE-Layout, SASA refers to the spatial-aware self-attention proposed by LayoutLMv2. $^\dag$ indicates the added module is proposed in this paper.} 
\label{tab:ablation_staudy}
\end{table*}

\begin{table}[ht]
\centering
\small
\begin{tabular}{llcccccc}
\toprule
\# & \textbf{Serialization Module} & \textbf{FUNSD} & \textbf{CORD} \\
\midrule
1 &\quad w/ Raster-Scan  & 0.9128  & 0.9658\\
2 &\quad w/ Layout-Parser    & 0.9143  & 0.9671 \\
3 &\quad w/ Document-Parser  & 0.9171  & 0.9678\\
\bottomrule
\end{tabular}
\caption{Performance analysis with different serialization modules, in which Raster-Scan means serialization with vanilla OCR results, while Layout-Parser and Document-Parser arrange the recognized words with the help of layout knowledge.}
\label{tab:layout_compare}
\end{table}

\noindent\textbf{Document Question Answering.}
Table~\ref{tab:main_result_docvqa} lists the Average Normalized Levenshtein Similarity (ANLS) score on DocVQA. 
Compared with all of the text-only baselines and best-performing multi-modal models, our method achieves competitive results and maximum performance improvement.
Note that LayoutLMv2(\#7) is developed based on UniLMv2(\#3), a model with powerful question-answering ability and even beat the multi-model model LayoutLM (\#4) on the task.
Unfortunately, UniLMv2 does not open any pre-training code or pre-trained model, and we can only use the parameters of RoBERTa to initialize our ERNIE-Layout. 
Nevertheless, we are surprised that ERNIE-Layout brings an exciting performance improvement to the backbone (almost double the increase of LayoutLMv2).
Furthermore, we achieve top-1 on the DocVQA leaderboard with ensemble.

\noindent\textbf{Document Image Classification.}
Table~\ref{tab:main_result_document_classification} shows the classification accuracy on RVL-CDIP, which again confirms the effectiveness of ERNIE-Layout in general document understanding.
Unlike these key information extraction or document question answering tasks focusing on multi-modal semantic understanding, document image classification requires a macro perception of text content and document layout.
Although our pre-training tasks pay attention to the fine-grained cross-modal matching, ERNIE-Layout still refreshes the best performance of the cross-grained task.

\subsection{Analysis}
We further conduct analysis experiments to study the effectiveness of the proposed pre-training tasks, attention mechanisms, and the serialization modules.
We select FUNSD and CORD as the evaluation datasets, keep all ablations sharing the same hyper-parameter settings, and report the average number of five runs with different random seeds.

\noindent\textbf{Effectiveness of Pre-training Tasks.} 
In this experiment, we start with the basic MVLM task to implement baseline models (\#1), and integrate new tasks step by step until the final model contains all four pre-training tasks (\#5).
From Table~\ref{tab:ablation_staudy}, we observe that RRP brings an improvement of 0.95\% on FUNSD, demonstrating the benefit of fine-grained cross-modal interaction.
When incorporating ROP, the performance of FUNSD is further increased by 1.3\%. We consider that ROP facilitates the model to learn a better representation that contains the reading order knowledge.

\noindent\textbf{Effectiveness of Attention Mechanisms.} 
LayoutLMv2~\cite{xu2021layoutlmv2} initially proposes spatial-aware self-attention to consider layout features in attention calculation, and many subsequent methods follow this idea.
From Table~\ref{tab:ablation_staudy}, we find that adopting such a mechanism can boost the performance of downstream tasks (\#4 v.s. \#6).
Meanwhile, disentangling attention into the position and content parts is another efficient solution to earn further performance gains (\#5 v.s. \#6).

\noindent\textbf{Effectiveness of Serialization Modules.}
Here we explore the impact of using different serialization modules on the downstream VrDU tasks.
As shown in Table~\ref{tab:layout_compare}, with the layout-knowledge based serialization modules (\#2, \#3), the model could achieve better performances (even without the disentangled attention).
We attribute the improvement to the fact that, although the advanced serialization is not used for fine-tuning datasets, the model has the ability to understand the proper reading order of documents after pre-training.

\section{Conclusion}

In this paper, we propose ERNIE-Layout, to integrate layout knowledge into document pre-training models from two aspects: serialization and attention.
ERNIE-Layout attempts to rearrange the recognized words of documents, which achieves considerable improvement on downstream tasks over the original raster-scanning order.
Besides, we also design a novel attention mechanism to help ERNIE-Layout build better interaction between text/image and layout features.
Extensive experiments demonstrate the effectiveness of ERNIE-Layout, and various analyses show the impact of different utilization of layout knowledge on VrDU tasks.

\section*{Acknowledgements}
This work was supported by Baidu Inc., the NSFC projects (No.~62072399), Chinese Knowledge Center for Engineering Sciences and Technology, and Artificial Intelligence Research Foundation of Baidu Inc., MoE Engineering Research Center of Digital Library.

\bibliography{custom}
\bibliographystyle{acl_natbib}

\appendix

\begin{table}
\centering
\small
\begin{tabular}{ c  c  c }
\toprule
\textbf{Document Page} &  \textbf{RS}  & \textbf{DP} \\ \midrule \midrule
    \begin{minipage}[b]{0.47\columnwidth}
        \centering
        \raisebox{-.5\height}{\includegraphics[width=\linewidth]{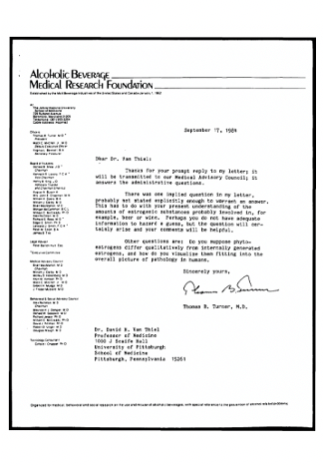}}
    \end{minipage}
    & 100.39
    & 67.98
    \\ \hline
    \begin{minipage}[b]{0.47\columnwidth}
        \centering
        \raisebox{-.5\height}{\includegraphics[width=\linewidth]{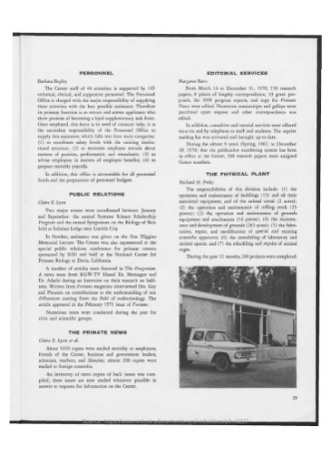}}
    \end{minipage}
    & 98.99
    & 42.02
    \\ \hline
    \begin{minipage}[b]{0.47\columnwidth}
        \centering
        \raisebox{-.5\height}{\includegraphics[width=\linewidth]{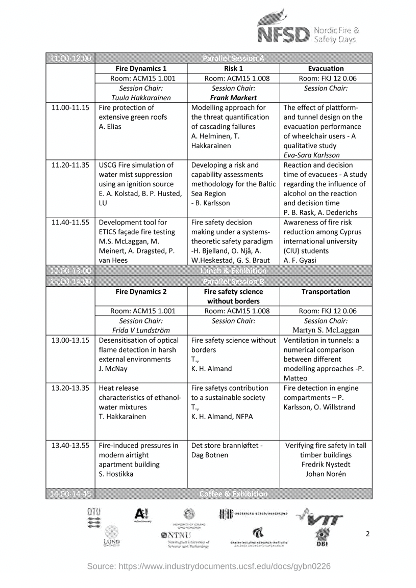}}
    \end{minipage}
    & 146.66
    & 76.87
    \\ \hline
    \begin{minipage}[b]{0.47\columnwidth}
        \centering
        \raisebox{-.5\height}{\includegraphics[width=\linewidth]{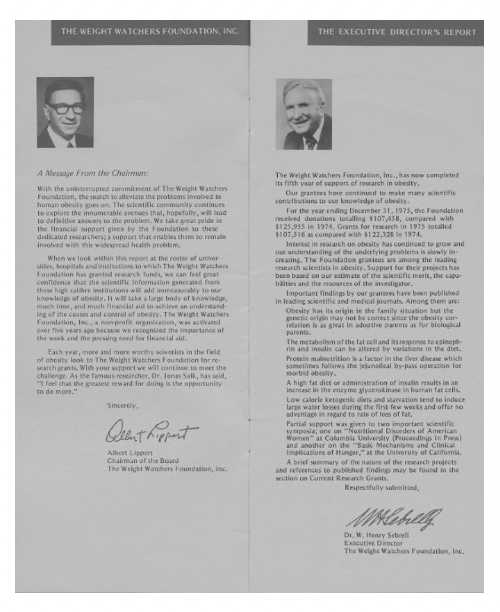}}
    \end{minipage}
    & 70.12
    & 25.61
    \\ \hline
    \begin{minipage}[b]{0.47\columnwidth}
        \centering
        \raisebox{-.5\height}{\includegraphics[width=\linewidth]{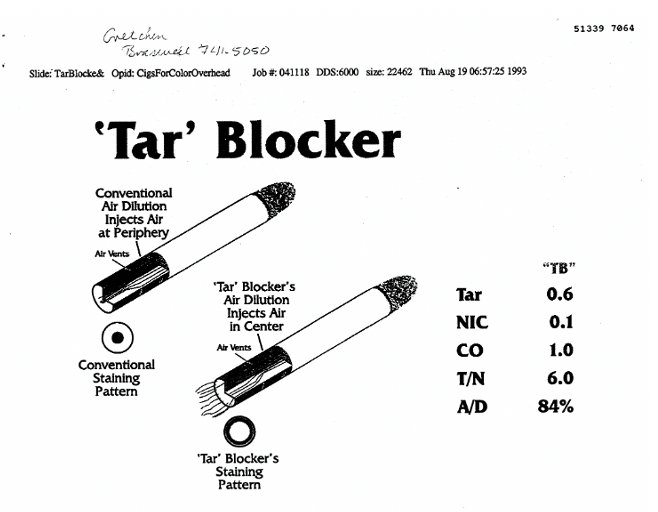}}
    \end{minipage}
    & 219.47
    & 170.54
    \\ \bottomrule
\end{tabular}
\caption{The PPL of serialized token sequence with different methods. RS refers to the Raster-scanning order and DP refers to the order with Document-Parser.}
\label{tab:document_parser}
\end{table}

\begin{figure*}
    \centering
    \includegraphics[width=0.7\textwidth]{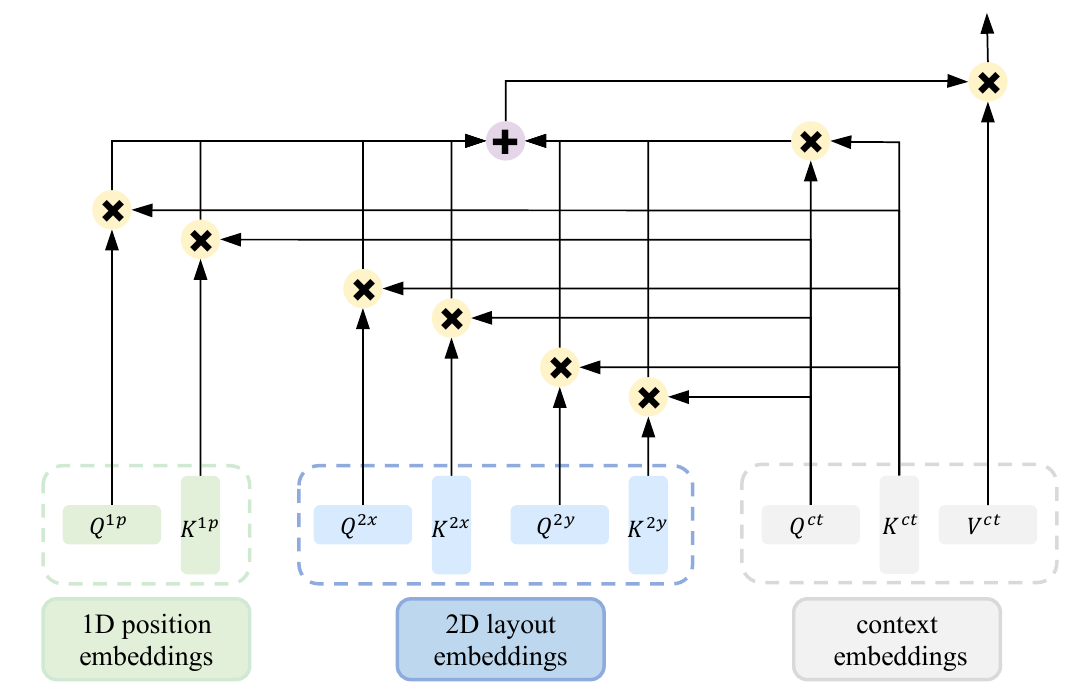}
    \caption{The internal working principle of spatial-aware disentangled attention.}
    \label{fig:disentange}
\end{figure*}

\section{Appendix}
\subsection{More Details about Document-Parser}
\label{sec:document_parser}

The Document-Parser assembles multiple modules such as document-specific OCR, Layout-Parser, and Table-Parser, in which the Layout Parser and Table Parser modules are crucial for incorporating layout knowledge in ERNIE-Layout.

An important preprocessing step for document understanding is serializing the extracted document tokens. The popular method for this serialization is performed directly on the output results of OCR in raster-scanning order and is sub-optimal though simple to implement. 
With the Layout-Parser and Table-Parser in the Document-Parser toolkit, the order of the tokens will be further rearranged according to the layout knowledge. During the parsing processing, the tables and figures are detected as spatial layouts, and the free texts are processed by paragraph analysis, combining heuristics and detection models to get the paragraph layout information and the upper-lower boundary relationship.

\begin{figure}[t]
    \includegraphics[width=\linewidth]{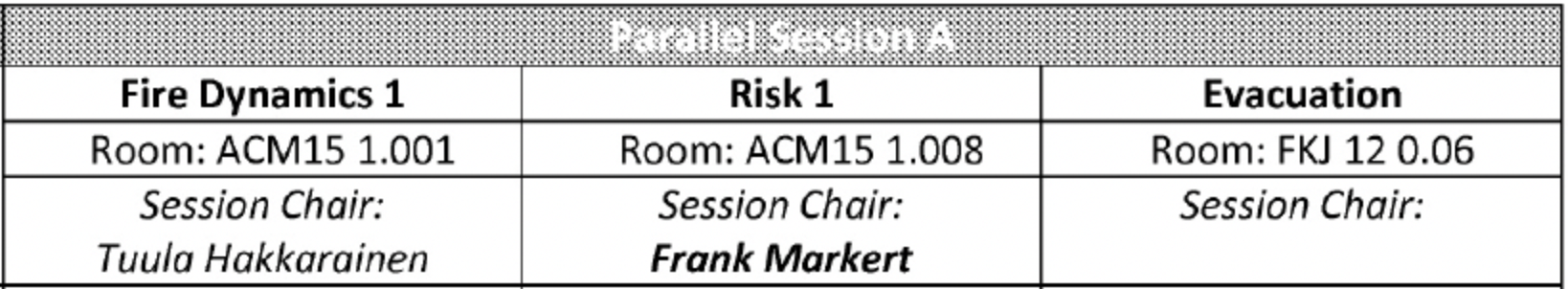}
    \caption{The example of a document with a complex layout. The serialization result with the raster-scanning order is ``\emph{... Session Chair: Session Chair: Session Chair: Tuula Hakkarainen ...}'', while serialization with Document-Parser is ``\emph{... Session Chair: Tuula wz Session Chair: Frank Markert ...}'', which is more consistent with human reading habits.}
    \label{fig:table_case}
\end{figure}

To validate the effectiveness of our method, we use an open-sourced language model GPT-2~\cite{radford2019language}, to calculate the PPL of the serialized token sequence by the raster-scanning order and Document-Parser respectively. Since documents with complex layouts only account for a small proportion of the total documents, in a test of 10,000 documents, the average PPL only drops about 1 point. However, for these documents with complex layouts, as shown in Table~\ref{tab:document_parser}, Document-Parser shows great advantages.
An example is shown in Figure \ref{fig:table_case}, which is extracted from the third image in Table~\ref{tab:document_parser}. to show the sequence serialized by Raster-Scan and Document-Parser.

\subsection{More Details about Multi-modal Transformer}
\label{sec:detail_disentangle}
Section~\ref{sec:muti-modal-transformer} describes the proposed spatial-aware disentangled attention for the multi-modal transformer through formulas.
To facilitate intuitive understanding, we also supplement the flow chart of calculation in Figure \ref{fig:disentange}.

\begin{table*}[t]
\centering
\small
\resizebox{\linewidth}{!}{
\begin{tabular}{lcccccc}
\toprule
\textbf{Methods} & \textbf{FUNSD} (F1) & \textbf{CORD} (F1) & \textbf{SROIE} (F1) & \textbf{Kleister-NDA} (F1) & \textbf{DocVQA} (ANLS) & RVL-CDIP (Acc) \\
\midrule
LayoutLM\textsubscript{base}~(\citeyear{xu2020layoutlm})            & 0.7866 & 0.9472 & 0.9438 & 0.8270 & 0.6979 & 0.9442 \\
TILT\textsubscript{base}~(\citeyear{powalski2021going})             & -      & 0.9511 & \textbf{0.9765} & -      & \textbf{0.8392}$^\dag$ & 0.9525 \\
LayoutLMv2\textsubscript{base}~(\citeyear{xu2021layoutlmv2})        & 0.8276 & 0.9495 & 0.9625 & \underline{0.8330} & 0.7808 & 0.9525 \\
DocFormer\textsubscript{base}~(\citeyear{appalaraju2021docformer})  & \underline{0.8334} & \underline{0.9633} & -      & -      & \underline{0.7878}  & \textbf{0.9617} \\
\midrule
\midrule
ERNIE-Layout\textsubscript{base} (ours)  & \textbf{0.9028} & \textbf{0.9661} & \underline{0.9719} & \textbf{0.8740} & 0.7758 & \underline{0.9581} \\ 
\bottomrule
\end{tabular}
}
\caption{Results of ERNIE-Layout (base-level model) and previous methods on various downstream VrDU tasks. $^\dag$ marks the results without any description of fine-tuning set (train or train+dev), The bold and underlined scores indicate the best and second results, respectively.}
\label{tab:base_results}
\end{table*}

\subsection{More Details about Experiments}
\label{sec:detail_finetuning_datasets}

\subsubsection{Finetuning Datasets}

\textbf{FUNSD}~\cite{jain2019multimodal} is a dataset for form understanding on noisy scanned documents that aims at extracting values from forms, which comprises 199 real, fully annotated, scanned forms. The training set contains 149 samples, and the test set contains 50 samples. We use the official OCR annotations. Following previous methods, we adopt entity-level F1 as the evaluation metric. Like StructralLM \cite{li2021structurallm}, we use the cell-level layout information when fine-tuning.

\textbf{CORD}~\cite{park2019cord} is a consolidated dataset for receipt parsing as the first step towards post-OCR parsing tasks. CORD consists of thousands of Indonesian receipts, including images, box/text annotations for OCR, and multi-level semantic labels for parsing. The training, validation, and test sets contain 800, 100, and 100 receipts, respectively. We use the official OCR annotations and entity-level F1 as the evaluation metric.

\textbf{SROIE}~\cite{huang2019icdar2019} is a scanned receipts OCR and key information extraction dataset, which covers important aspects related to the analysis of scanned receipts. The training and test set contain 626 and 347 samples, respectively. This task requires the model to extract values from each receipt of four predefined keys: company, date, address, and total. We use the official OCR annotations and entity-level F1 as the evaluation metric. 

\textbf{Kleister-NDA}~\cite{gralinski2021kleister} is provided for key information extraction task, which involves a mix of scanned and born-digital long formal documents. The training, valid, and test sets contain 254, 83, and 203 samples, respectively. Due to the test set is not publicly available, we report the entity-level F1 score on the validation set, which is computed by the official evaluation tools\footnote{https://gitlab.com/filipg/geval}. The task aims to extract values of four predefined keys: date, jurisdiction, party, and term. 

\textbf{RVL-CDIP}~\cite{harley2015evaluation} is a document classification dataset consisting of grayscale document images. The training, validation, and test sets contain 320000, 40000, and 40000 document images, respectively. The document images are categorized into 16 classes, with 25000 images per class. We use Microsoft OCR tools to extract text and layout information from document images, and the evaluation metric is classification accuracy. 

\textbf{DocVQA}~\cite{mathew2021docvqa} is a dataset for Visual Question Answering (VQA) on document images. The dataset consists of 50000 questions defined on 12767 document images. The document images are split into the training set, validation set, and test set with the ratio of 8:1:1. We use the Microsoft OCR tools to extract the texts and layouts from document images. 
The task aims to predict the start and end position of the answer span. Average Normalized Levenshtein Similarity~\cite{biten2019icdar} is used as the evaluation metric. 

\subsubsection{Baselines}

In the experiment, we compare ERNIE-Layout with two groups of recent models: \textbf{text-only models} (BERT~\cite{devlin2019bert}, RoBERTa~\cite{liu2019roberta}, UniLMv2~\cite{bao2020unilmv2}) and \textbf{multi-modal models} (LayoutLM~\cite{xu2020layoutlm}, LayoutLMv2~\cite{xu2021layoutlmv2}, TILT~\cite{powalski2021going}, StructuralLM~\cite{li2021structext}, DocFormer~\cite{appalaraju2021docformer}).
Note that LayoutLM is initialized from BERT, LayoutLMv2 is initialized from UniLMv2, TILT is initialized from T5,  StructuralLM and our ERNIE-Layout are initialized from RoBERTa.

\subsubsection{Results with RoBERTa-base}

In the main content, we leverage RoBERTa-large to initialize ERNIE-Layout and compare it with the previous same-level model. 
Here, we also compare the base-level model with 12 transformer layers (768 hidden state and 12 attention heads), that is, initializing ERNIE-Layout with RoBERTa-base.
We omit StructuralLM since it does not release the parameters and performances of its base model.
From the results in Table~\ref{tab:base_results}, it is easy to observe a similar phenomenon with ERNIE-Layout\textsubscript{large}: ERNIE-Layout\textsubscript{base} also achieves significant performance improvement on various VrDU tasks, especially in FUNSD and Kleister-NDA, but slightly poor in DocVQA (detailed analysis and further exploration have been given in Section~\ref{sec:exp_results}).
By the way, We are also pleasantly surprised to find that ERNIE-Layout\textsubscript{base} even beats some large-level model in kinds of datasets (e.g., FUNSD, CORD, Kleister-NDA, RVL-CDIP).

\end{document}